\documentclass[preprint,12pt]{elsarticle}

\usepackage{amssymb}
\usepackage{amsmath}
\usepackage[hidelinks]{hyperref}
\usepackage{caption,tabularx,booktabs, array, makecell, subcaption}
\usepackage{appendix}
\usepackage{orcidlink}
\newcommand{\PreserveBackslash}[1]{\let\temp=\\#1\let\\=\temp}
\newcolumntype{C}[1]{>{\PreserveBackslash\centering}p{#1}}
\newcolumntype{R}[1]{>{\PreserveBackslash\raggedleft}p{#1}}
\newcolumntype{L}[1]{>{\PreserveBackslash\raggedright}p{#1}}

\journal{}

\begin{document}

\begin{frontmatter}



\title{Can Reasoning LLMs Enhance Clinical Document Classification?\tnoteref{Can Reasoning LLMs Enhance Clinical Document Classification?}}
% \tnotetext[Human vs Large Language Models for Classifying Clinical Documents]{}
 
\author[label1]{Akram Mustafa\orcidlink{0000-0003-4090-2597}}
\ead{akram.mohdmustafa@my.jcu.edu.au}

\author[label2]{Usman Naseem\orcidlink{0000-0003-0191-7171}}
\ead{usman.naseem@mq.edu.au}

\author[label1]{Mostafa Rahimi Azghadi\orcidlink{0000-0001-7975-3985}}
\ead{mostafa.rahimiazghadi@jcu.edu.au}
\cortext[cor1]{mostafa.rahimiazghadi@jcu.edu.au}

 
 \affiliation[label1]{organization={College of Science and Engineering, James Cook University},
             city={Townsville},
             postcode={4811},
             state={QLD},
             country={Australia}}


 \affiliation[label2]{organization={School of Computing, Macquarie University},
             city={Sydney},
             postcode={2113},
             state={NSW},
             country={Australia}}


%% Abstract
\begin{abstract}
\noindent\textit{Background}: Clinical document classification is a critical process in healthcare, converting unstructured medical texts into standardized ICD-10 diagnoses. This process faces challenges due to the complex and varied nature of medical language, which includes domain specific terminology, abbreviations, and unique writing styles across institutions. Additionally, privacy regulations and limited high quality annotated datasets hinder the development of robust models. LLMs have emerged as a transformative technology in healthcare, improving the efficiency and accuracy of tasks like clinical document classification by leveraging advanced natural language understanding. \\

\noindent\textit{Objective}: The objective of this study is to evaluate the performance and consistency of LLMs in classifying clinical discharge summaries based on ICD-10 codes. By leveraging both reasoning and non-reasoning LLMs, the study aims to determine how effectively these models can identify and classify clinical patterns, providing insights into their potential for improving automated clinical coding accuracy and enhancing decision support in healthcare settings. \\

\noindent\textit{Methods}: This study used a balanced subset of the MIMIC-IV dataset, comprising 3,000 discharge summaries including 150 positive and 150 negative samples for each of the top 10 ICD-10 codes.
The summaries were tokenized using cTAKES, which converted clinical narratives into structured SNOMED codes, capturing contextual details such as affirmation or negation. Eight LLMs, including four reasoning (Qwen QWQ, Deepseek Reasoner, GPT o3 Mini, Gemini 2.0 Flash Thinking) and four non-reasoning models (Llama 3.3, GPT 4o Mini, Gemini 2.0 Flash, Deepseek Chat), were evaluated over three experimental runs. Final predictions were determined using majority voting across the runs to assess accuracy, F1 score, and consistency.\\

\noindent\textit{Results}: Among the eight evaluated LLMs, reasoning models demonstrated superior performance in ICD-10 classification, achieving an average accuracy of 71\% and an F1 score of 67\%, compared to 68\% accuracy and 60\% F1 score for non-reasoning models. Gemini 2.0 Flash Thinking achieved the highest accuracy at 75\% and F1 score at 76\%, while GPT 4o Mini had the lowest performance 64\% accuracy, and 47\% F1 score. Consistency analysis revealed that non-reasoning models exhibited higher stability of 91\% average consistency than reasoning models of 84\%. Performance variations across ICD-10 codes highlighted strengths in identifying well defined conditions but challenges in classifying abstract diagnostic categories. \\

\noindent\textit{Conclusion}: 
The evaluation of reasoning and non-reasoning LLMs in ICD-10 classification highlights a trade-off between accuracy and consistency. Reasoning models achieved higher classification accuracy and F1 scores, excelling in complex clinical cases, while non-reasoning models demonstrated superior stability across repeated trials. These findings suggest that a hybrid approach, leveraging the strengths of both model types, could optimize automated clinical coding by balancing accuracy and reliability. Future research should explore multi-label classification, domain specific fine tuning, and ensemble modeling to enhance performance and generalizability in real-world healthcare applications.\\

\end{abstract}


\end{frontmatter}

\section{Introduction}\label{sec_Introduction}
Clinical document classification faces significant challenges due to the complex nature of medical narratives. The global clinical coding market size was valued at more than \$37 Billion USD in 2024 \cite{963}. The clinical coding process faces significant challenges due to the complex, varied nature of medical narratives. These documents use heavy domain specific terminologies, abbreviations, and unique writing styles that vary by institution and practitioner. 
Moreover, in addition to healthcare strict privacy regulations, the limited availability of large, high quality annotated datasets slows down the development of robust classification models. The frequent imbalance in class distributions and the inherent ambiguity in clinical language further complicate model training and evaluation. To address these issues, specialized Natural Language Processing (NLP) frameworks such as the clinical Text Analysis and Knowledge Extraction System (cTAKES) have been developed to better handle the unique characteristics of clinical texts \cite{390}.

Clinical document classification is one of the processes that the healthcare industry relies on to convert unstructured medical documents to International Classification of Diseases, 10th Revision (ICD-10) diagnoses \cite{965}. ICD-10 coding has become an important process of clinical practice by providing a standardized language for documenting diagnoses and procedures. This systematic classification facilitates efficient communication among healthcare providers, ensuring that patient information is accurately recorded and easily shared across different institutions and systems \cite{977,978}. The consistent application of ICD-10 codes not only enhances clinical documentation and billing processes but also plays a pivotal role in epidemiological research, quality assurance, and public health reporting \cite{964}. By enabling comprehensive tracking of disease trends and treatment outcomes, ICD-10 coding informs health policy and resource allocation, ultimately contributing to improved patient care and safety \cite{979,980}.

Large language models (LLMs) have rapidly emerged as a transformative technology in healthcare, driven by their advanced natural language understanding and reasoning capabilities \cite{981}. These models, trained on extensive and diverse corpora that include medical literature and clinical narratives, can parse complex clinical language, infer relationships among diverse medical concepts, and generate contextually nuanced insights \cite{966, 967}. Such reasoning abilities are particularly beneficial for automating tasks like clinical document classification and diagnosis coding, where they can support decision-making and improve the efficiency and accuracy of healthcare delivery. Recent developments in LLM architectures, including techniques like chain-of-thought processing, have further enhanced their capacity to handle the intricacies of clinical data, marking a significant shift from traditional NLP approaches \cite{968, 969}.

Comparing reasoning versus non-reasoning large language models addresses a crucial research gap in applying these systems to complex clinical tasks. While models enhanced with explicit reasoning have shown promise in capturing intricate medical relationships and nuances in clinical narratives \cite{970, 971}, their benefits over standard non-reasoning models have not been systematically quantified in clinical document classification. This comparison is essential to determine whether the additional computational complexity and resource demands of reasoning-based approaches translate into significantly improved diagnostic coding accuracy and consistency, ultimately informing more effective and interpretable model deployment in healthcare settings.

This study addresses several critical research questions that bridge the gap between reasoning and non-reasoning LLMs in clinical document classification. First, it investigates how reasoning LLMs compare with their non-reasoning counterparts in classifying ICD-10 coded discharge summaries, exploring whether the enhanced reasoning abilities translate to improved handling of clinical narratives. Second, the research evaluates the overall classification accuracy of LLMs on tokenized discharge summaries, providing quantitative insights into their performance on a challenging and balanced dataset. Third, the study examines the consistency of classification results across multiple experimental runs for each model, thereby assessing the robustness and reliability of these approaches. Collectively, these research questions aim to illuminate the strengths and limitations of current LLM methodologies in clinical document classification, offering valuable contributions toward the optimal deployment of AI in clinical settings.

\section{Related Work}\label{sec_RelatedWork}
\subsection{Clinical Text Classification}
Recent advances in clinical text classification have progressively shifted from traditional machine learning techniques to more nuanced deep learning methods that better capture the complexity of clinical language. For instance, the work by Haoran Shi et al. \cite{377} introduces a hierarchical deep learning model with an attention mechanism specifically designed for the automated assignment of ICD diagnostic codes from written diagnosis descriptions. Their approach leverages character-aware neural language models to generate detailed hidden representations of both clinical narratives and ICD codes, effectively addressing the inherent mismatch between the number of descriptions and their corresponding codes. The significant performance gains, marked by an F1 score of 53\% and an AUC of 90\%, underscore the potential of deep learning frameworks over conventional methods that often rely on manual feature engineering.

Gehrmann et al. \cite{433} explored the potential of deep learning for patient phenotyping which is a clinical text classification task closely related to automated ICD coding, by leveraging discharge summaries from the MIMIC-III dataset. Their study compared convolutional neural networks (CNNs) with traditional n-gram models and cTAKES based approaches, demonstrating that CNNs could automatically learn salient textual features, thereby significantly outperforming the other methods across 10 distinct phenotyping tasks. With an average F1 score of 76\%, their findings highlight the advantages of deep learning in capturing complex clinical information, reducing reliance on manually crafted annotation rules, and improving interpretability. While the focus of their work was on patient phenotyping, the methodological insights and demonstrated performance gains underscore the relevance of deep neural architectures for ICD coding tasks as well, further cementing the MIMIC dataset’s role as a critical benchmark in clinical NLP research.

\subsection{Tokenization \& Preprocessing in Clinical NLP}\label{sec_Token}
In clinical NLP, the preprocessing of unstructured medical texts is essential for ensuring that downstream models can effectively interpret clinical narratives. In his work on classifying medical notes into standard disease codes \cite{426}, Karmakar emphasizes a systematic approach to note preprocessing that includes standard steps such as lowercasing, removal of special characters, and tokenization. Tools like cTAKES are particularly relevant in this context because they provide domain-specific processing capabilities tailored to the nuances of clinical language. cTAKES not only streamlines tokenization and sentence boundary detection but also offers robust named entity recognition that can identify medical terms, abbreviations, and concepts despite variations in clinical documentation. This targeted preprocessing is crucial for managing the inherent variability of discharge summaries and ultimately contributes to the effectiveness of models, such as CNNs and Long Short Term Memory models (LSTMs), when classifying complex clinical data into ICD codes.

\subsection{Large Language Models for Clinical Coding}\label{sec_LLM}
Li \emph{et al.}'s work \cite{958} provides a study exploring the use of multiple LLM agents collaborating to assign ICD codes to clinical notes. In this framework, distinct agents were tasked with reading clinical notes, generating relevant ICD codes, and then engaging in a discussion to reach a consensus. By structuring the interaction as a multi-agent dialogue, the authors aimed to simulate expert deliberation and improve coding accuracy. Their findings demonstrated that this collaborative LLM approach significantly enhanced performance, highlighting the potential of agent-based coordination in complex medical classification tasks..

Mustafa \emph{et al.}'s work \cite{922} is closely aligned with the ongoing exploration of LLMs for clinical coding. By assessing ChatGPT 3.5 and 4 compared to traditional machine learning and SNOMED mapping methods, their study investigates the challenge of accurately classifying ICD-10 codes in complex medical records. They demonstrate that, while human coders still achieve superior accuracy, ChatGPT 4’s ability to match the median human performance and exhibit improved consistency suggests that advanced LLMs can play a valuable role in clinical coding. 

These studies highlight both the potential and challenges of using LLMs for clinical coding. While they show promising performance, issues such as hallucination and handling complex imbalanced code sets remain. Approaches that incorporate reasoning, domain knowledge, and structured workflows offer a path forward. In this research, we investigate if reasoning LLMs can enhance document classification.

\section{Materials and Methods}\label{sec_Method}
\subsection{Dataset Preparation}\label{sec_Prep}
The MIMIC IV dataset, which contains a diverse range of discharge summaries annotated with ICD-10 diagnoses, was used. These diagnoses were classified into two clinically significant categories: the Top 5 Primary Diagnoses, directly extracted from the discharge summaries and representing the main clinical focus of each case, and the Top 5 Secondary Diagnoses, capturing additional clinical details that, while not the primary focus, are important for patient care and outcomes (see Table \ref{table_FreqDiag}).

The analysis incorporated both primary and secondary diagnoses, forming a set of the top 10 ICD-10 codes. This dual list approach was selected to capture a more comprehensive spectrum of clinically relevant conditions. By combining both primary and secondary diagnoses, the study benefits from a richer diagnostic landscape, which is essential for robust evaluation of classification performance across diverse clinical scenarios. To ensure class balance within the dataset, for each of these 10 ICD-10 codes, 150 discharge summaries confirmed as positive (i.e., diagnosed with the respective ICD-10 code) were extracted. An equal number of 150 negative discharge summaries, not associated with the given ICD-10 code, were randomly sampled from the set of ICD codes, excluding the 10 selected codes. This approach yielded a total of 300 discharge summaries per ICD-10 code, resulting in an aggregate dataset of 3,000 discharge summaries.

The balanced nature of this dataset, along with the inclusion of both primary and secondary diagnoses, forms the foundation for the subsequent experiments evaluating the performance and consistency of various large language models in the classification task. It is important to note that due to the significant computational and financial costs associated with running multiple large language models, the dataset was intentionally limited to 3,000 discharge summaries. This decision ensured that the experiments remained feasible within the available resources while still providing a robust evaluation framework for the classification tasks.

\begin{table}[t]

%\begin{table}
\caption{Top 10 ICD codes used in this study. The top part of the table shows 5 codes with the highest number of primary diagnoses cases, while the bottom part shows the top 5 codes with the maximum total cases in the MIMIC IV dataset.}
\label{table_FreqDiag}
\smallskip

\begin{tabular}{@{} L{190pt} C{40pt}C{40pt}C{85pt} @{}}
\toprule
			Diagnosis
			&\makecell{ICD-10\\Code}
                &\makecell{Total\\Cases}
			&\makecell{Primary\\Diagnosis Cases} \\

\midrule
		
			Sepsis	&A41	&7,430 &4,830	\\ 			
			Myocardial infarction	&I21	&5,735  &2,722	\\ 			
			Other medical care  &Z51	&6,919    &2,370	\\ 			
			Chronic ischaemic heart disease	&I25	&38,157  &2,302	\\ 			
			Hypertensive heart and renal disease    &I13 &8,366    &2,114	\\ 			
\midrule

			Disorders of lipoprotein metabolism and other lipidemias	&E78	&49,310 &9	\\ 			
			Essential hypertension	&I10	&43,574  &76	\\ 			
			Long term drug therapy	&Z79	&40,393  &0	\\ 						Personal history of other diseases and conditions    &Z87 &40,255    &0	\\ 			

			Place of occurrence of the external cause  &Y92	&35,297    &0	\\ 			
\bottomrule
\end{tabular}
%\end{table}

\end{table}

\subsection{Preprocessing}\label{sec_PreProc}
cTAKES was utilized to convert unstructured discharge summaries into structured, machine-interpretable data by transforming the clinical narratives into lists of standardized SNOMED codes. These codes represent a comprehensive array of clinical entities such as diseases, lab tests, procedures, medications, symptoms, anatomy, and events. During the tokenization process, cTAKES not only extracted these terms but also determined their contextual status categorizing each as either affirmed or negated. This automated pipeline significantly reduced the length of the original documents by eliminating unnecessary text, which in turn improved downstream processing efficiency and cost. After processing, the extracted SNOMED items were aggregated into two separate sections based on their contextual designation, and the frequency of each term's occurrence was recorded. This comprehensive approach provided a robust dataset of tokenized discharge summaries, enriched with detailed frequency counts and contextual annotations, which served as the foundation for subsequent classification experiments using large language models.

\subsection{Experimental Setup}\label{sec_Experiment}
A total of eight large language models were chosen to evaluate the classification of clinical discharge summaries based on ICD-10 codes. The models were categorized into two groups: reasoning and non-reasoning, as shown in Table \ref{table_LLMs}. 
The deliberate selection of these eight models enabled a comprehensive comparison between architectures that incorporate explicit reasoning and those that do not, thereby addressing the research question concerning the impact of reasoning on clinical document classification.

\begin{table}[!t]
\caption{Large Languages Models used.}
\label{table_LLMs}
\begin{tabular}{@{}L{180pt}C{100pt}C{90pt} @{}}
\toprule

\textbf{\makecell{Model}}
&\textbf{\makecell{Reasoning Type}}
&\textbf{\makecell{Platform}} \\

\midrule
		
Deepseek Reasoner	        &Reasoning	&Deepseek \\
Gemini 2.0 Flash Thinking	&Reasoning	&Google \\
GPT o3 Mini                 &Reasoning	&OpenAI \\
Qwen QWQ 32B                &Reasoning	&Groq \\
\midrule
Deepseek Chat   	        &Non-Reasoning	&Deepseek \\
Gemini 2.0 Flash        	&Non-Reasoning	&Google \\
GPT 4o Mini                 &Non-Reasoning	&OpenAI \\
Llama 3.3 70B Versatile     &Non-Reasoning	&Groq \\

\bottomrule
\end{tabular}

\end{table}
For the experimental procedure, each of the eight models was tasked with classifying a dataset of 3,000 tokenized discharge summaries that had been preprocessed using cTAKES. To ensure the robustness and repeatability of the results, each model was run three times, with each run independently processing the entire dataset. In every experimental iteration, the models generated classifications indicating whether each discharge summary belonged to its corresponding ICD-10 group. The prompt used for all models was: 

\textit {``Discharge Summary: [Report Text]  Does this summary contain the diagnosis associated with ICD-10 code [ICD10 Code] or any of its specific subcategories, Answer with Yes or No only?"}

\begin{figure}[t]
\centering
\includegraphics[width=1.0\textwidth]{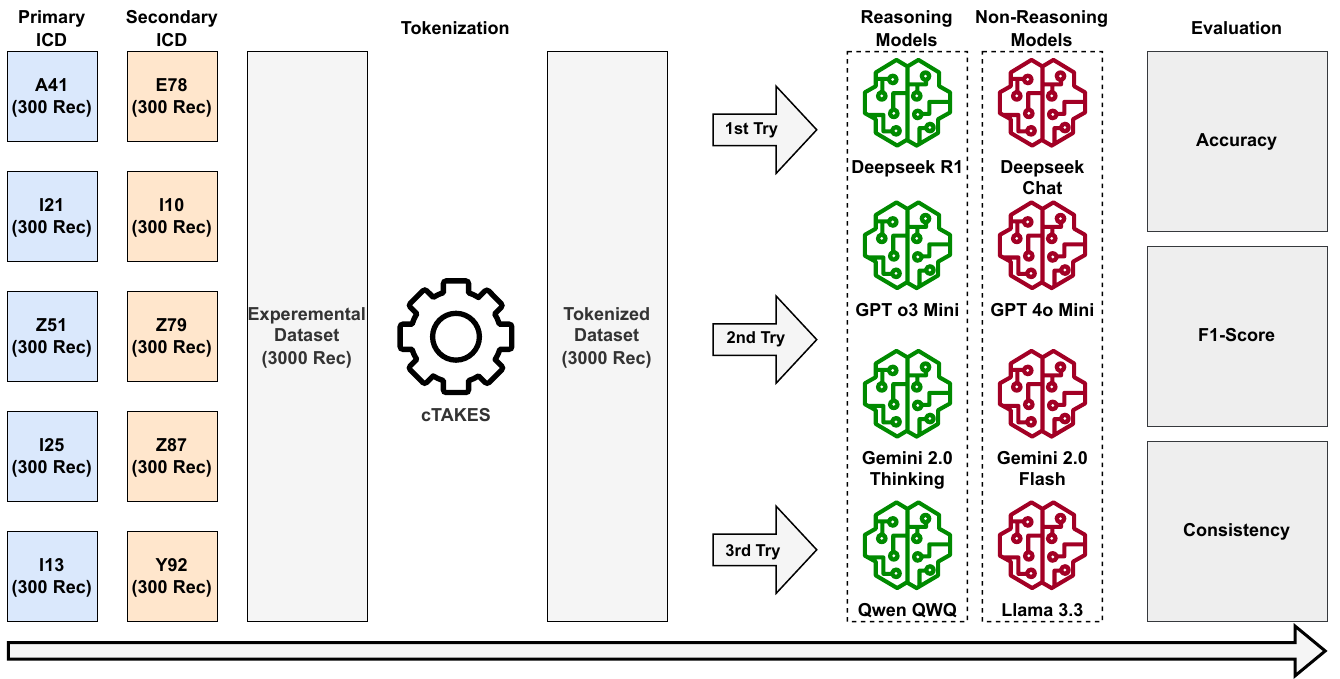}
\noindent
\caption{Workflow of cTAKES based Tokenization and Repeated LLM Classification for ICD-10 coded Discharge Summaries.}
\label{graph_process}
\end{figure}

The final classification outcome for each summary was determined using a majority vote across the three repetitions, meaning that the label assigned by at least two out of three runs was taken as the definitive result. This methodological approach not only ensured consistency in the classification outcomes but also provided a reliable measure of each model's performance in distinguishing between the ICD-10 groups. Figure \ref{graph_process} illustrates this process. In addition, in favour of reproducible research, our code and data are shared publicly on GitHub: \href{https://github.com/asmgx/LLMs}{https://github.com/asmgx/LLMs}.

\subsection{Evaluation Metrics}\label{sec_Metric}
Once final classifications were established, overall accuracy was computed to evaluate the models. The F1 score provided a more robust assessment of classification accuracy than accuracy alone. Together, these metrics offered deeper insight into the models' predictive reliability.

Lastly, we assessed how consistently each model classified the same discharge summaries across its three runs. Specifically, for each discharge summary, we recorded whether the model’s prediction (positive or negative) remained the same in all three trials. The proportion of summaries that received identical classifications in all runs was then computed as an indicator of consistency. This measure offers insight into the reliability of the model’s decision-making process, revealing whether fluctuations in output are minimal (high consistency) or more pronounced (low consistency) over repeated evaluations.

\section{Results}\label{Sec_Results}
\subsection{LLMs Performance}\label{Sec_PerformanceMetrics}
Among the eight LLMs evaluated, Gemini 2.0 Flash Thinking, a reasoning model, achieved the highest overall accuracy at 75\% and the highest F1 score at 76\%. In contrast, GPT 4o Mini, a non-reasoning model, had the lowest performance among all models, with an accuracy of 64\% and an F1 score of 47\%. Among the non-reasoning models, Gemini 2.0 Flash recorded the highest accuracy at 72\% and the highest F1 score at 71\%, while Qwen QWQ had the lowest accuracy and F1 score among reasoning models, with 65\% accuracy and 55\% F1 score. Table \ref{table_Models} presents the detailed accuracy and F1 scores for each model.

\begin{table}[h]
\caption{Average performance of various LLMs over 3 runs in classifying 3000 clinical discharge summaries from the top 10 ICD-10 codes within the MIMIC IV dataset.}
\label{table_Models}
\begin{tabular}{@{}L{150pt}C{60pt}C{60pt}C{80pt} @{}}
\toprule

\textbf{\makecell{Model}}
&\textbf{\makecell{Accuracy}}
&\textbf{\makecell{F1 Score}}
&\textbf{\makecell{Consistency}} \\

\midrule
		
Deepseek Reasoner			&73.83\%	&69.91\%	&82.83\% \\
Gemini 2.0 Flash Thinking	&\textbf{75.30\%}	&\textbf{75.50\%}	&78.43\% \\
GPT 3o Mini					&71.37\%	&63.59\%	&\textbf{95.47\%} \\
Qwen QWQ 32B				&65.23\%	&54.83\%	&78.07\% \\
\midrule
Deepseek Chat				&69.80\%	&59.48\%	&95.40\% \\
Gemini 2.0 Flash			&71.93\%	&70.50\%	&90.67\% \\
GPT 4o Mini					&63.77\%	&46.69\%	&88.73\% \\
Llama 3.3 70B Versatile		&68.33\%	&60.15\%	&88.30\% \\

\bottomrule
\end{tabular}
\end{table}

\begin{figure}[t]
\centering
\includegraphics[width=1.0\textwidth]{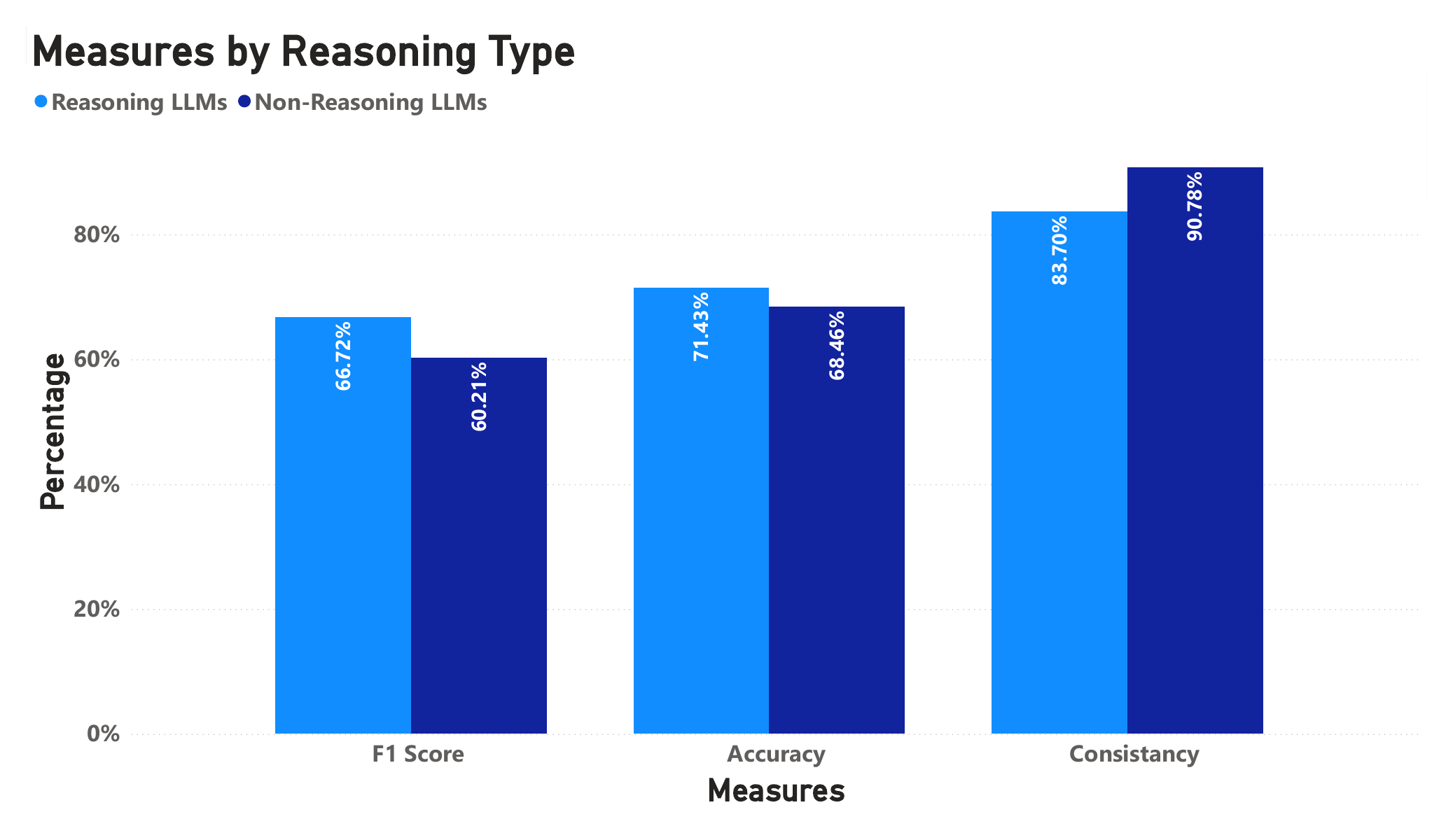}
\noindent
\caption{Average performance of 4 reasoning vs 4 non-reasoning LLMs in classifying clinical documents.}
\label{g_Measure_Reason}
\end{figure}

Overall, the reasoning models outperformed the non-reasoning models in terms of average accuracy and F1 scores. The reasoning models achieved an average accuracy of 71\% compared to 68\% for the non-reasoning models. Similarly, the reasoning models recorded an average macro F1 score of 67\%, whereas the non-reasoning models achieved a lower average macro F1 score of 60\%. Figure \ref{g_Measure_Reason} shows the average performance of the four state-of-the-art reasoning versus four non-reasoning LLMs in classifying clinical documents. These results suggest that, in general, the reasoning models provided slightly better classification performance than the non-reasoning models.

\subsection{Consistency Analysis}\label{Sec_ConsistencyAnalysis}
The consistency of model performance across the three experiments reveals a varied level of reliability among the different LLMs. Results are shown in Table \ref{table_Models}. Models such as  GPT o3 Mini and Deepseek Chat at 95\%, and Gemini 2 Flash at 91\% demonstrate high consistency, with performance largely stable across all three trials. This suggests that these models are robust in classifying discharge summaries, maintaining similar results across multiple runs. On the other hand, models like Qwen QWQ and Gemini 2 Flash Thinking, with a consistency rate of 78\%, show significantly lower consistency, indicating greater variability in their performance across the experiments. This variability could potentially indicate issues with stability or sensitivity to different input data or model configurations. Other models such as Deepseek Reasoner at 83\%, Llama 3.3 at 88\%, and GPT 4o Mini at 89\% fall in between, suggesting a moderate level of consistency. Overall, the high consistency observed in certain models like Deepseek Chat and GPT o3 Mini suggests they are more reliable for repeated use in clinical document classification tasks. 
An analysis of the average consistency based on model reasoning, reveals that non-reasoning models exhibit a higher average consistency of 91\% compared to reasoning models, which show an average consistency of 84\%. Figure \ref{g_Measure_Reason} illustrates these findings in detail.

\subsection{Detailed Performance Analysis of Various LLMs}
The performance analysis of the eight investigated LLMs in classifying ICD-10 codes is presented in two different ways. Tables \ref{table_A41} to \ref{table_Z87} list the accuracy, F1 score, and consistency of all LLMs on each specific code. Additionally, Figures \ref{g_Rad_A41} to \ref{g_Rad_Z87} demonstrate the F1 score of all LLMs vs their consistency, for each code. These results assist in classifying the codes into three categories.

The first category includes codes for which all models demonstrate strong performance in both consistency and F1 scores, as shown in Figures \ref{g_Rad_A41}, \ref{g_Rad_i10}, \ref{g_Rad_i21}, and \ref{g_Rad_i25}, exhibiting similar trends and levels of accuracy. This category comprises ICD-10 codes A41, I10, I21, and I25. For these codes, consistency ranges from 67\% to 100\%, while F1 scores range from 63\% to 96\%, as shown in Tables \ref{table_A41}, \ref{table_I10}, \ref{table_I21}, and \ref{table_I25}. Reasoning models, particularly GPT 3o Mini, stood out in terms of consistency, achieving the highest levels for 3 out of the 4 codes. In terms of F1 scores, both reasoning and non-reasoning models led, depending on the specific ICD code. Notably, GPT 4o Mini achieved the highest F1 score of 96\%.

The second category includes codes where consistency remained high and stable across all models, but F1 scores varied significantly, as shown in Figures \ref{g_Rad_E78}, \ref{g_Rad_i13}, \ref{g_Rad_Y92}, and \ref{g_Rad_Z51}. This suggests that while the models consistently made similar predictions, their accuracy in identifying true positive cases varied widely. ICD-10 codes in this category include E78, I13, Y92, and Z51. Consistency scores ranged from 64\% to 100\%, while F1 scores showed a much wider spread—ranging from 23\% to 79\% for E78, 30\% to 86\% for I13, 0\% to 20\% for Y92, and 4\% to 87\% for Z51, as shown in Tables \ref{table_E78}, \ref{table_I13}, \ref{table_Y92}, and \ref{table_Z51}. GPT 4o Mini showed notably poor performance in this category, recording the lowest F1 scores for E78, I13, and Y92 at 23\%, 30\%, and 0\%, respectively.

The third category includes codes Z79 and Z87, where both consistency and F1 scores varied greatly across models, indicating inconsistent performance in identifying positive discharge summaries, as shown in Figures as shown in Figures \ref{g_Rad_Z79}, \ref{g_Rad_Z87}. For Z79, consistency ranged from 33\% to 100\%, while F1 scores ranged from 0\% to 57\%. For Z87, consistency ranged from 36\% to 100\%, and F1 scores ranged from 0\% to 62\%, as detailed in Tables \ref{table_Z79} and \ref{table_Z87}. GPT 3o Mini demonstrated 100\% consistency for both codes but scored 0\% on F1, indicating that it consistently misclassified all records as negative. Similarly, Deepseek Chat and GPT 4o Mini also failed to identify any positive cases for Z87, both recording an F1 score of 0\%, although their consistency levels were slightly lower.

\section{Discussion}\label{Sec_Discussion}
\subsection{Interpretation of Findings}\label{Sec_Interpretation}
The findings of this study indicate that reasoning  LLMs generally outperform non-reasoning models in ICD-10 classification tasks, achieving higher accuracy and F1 scores on average. This aligns with an initial hypothesis that explicit reasoning capabilities would enhance the ability of LLMs to capture complex clinical relationships within discharge summaries. However, the results also reveal unexpected patterns. While reasoning models demonstrated superior classification performance, they exhibited lower consistency across repeated experiments than non-reasoning models. This suggests that while reasoning models may improve accuracy, they could also introduce greater variability in results, possibly due to sensitivity to input variations or internal model heuristics. Additionally, while some reasoning LLMs excelled in identifying well-defined medical conditions such as sepsis and myocardial infarction, they struggled with more abstract ICD-10 codes related to patient history and external causes, reinforcing concerns about their contextual understanding. These findings highlight both the potential and limitations of reasoning LLMs, underscoring the need for further refinement to enhance their reliability in clinical applications.

\subsection{Clinical Relevance}\label{Sec_ClinicalRelevance}
The optimization of ICD-10 coding workflows and the broader integration of AI in clinical documentation are of significant clinical relevance. The superior accuracy of reasoning LLMs suggests that they could serve as valuable tools for assisting clinical coders in accurately classifying discharge summaries, potentially reducing manual workload and minimizing coding errors. However, the observed variability in their consistency raises concerns about reliability, highlighting the importance of model calibration and validation before deployment in real world settings. Additionally, the inability of some models to correctly classify abstract or context-dependent ICD-10 codes underscores the need for further refinements, such as enhanced domain-specific training and better handling of nuanced medical contexts. If properly fine-tuned and integrated, these AI-driven coding assistants could streamline the ICD-10 classification process, improve coding accuracy, and ultimately support more efficient clinical documentation and billing workflows, leading to better healthcare data quality and decision making.

\subsection{Future Work}\label{Sec_FutureWork}
Future research could expand the scope of ICD-10 classification by incorporating all ICD-10 codes mentioned in discharge summaries, including both primary and secondary diagnoses. This would provide a more comprehensive evaluation of model performance across a broader range of clinical scenarios, capturing complex multi-diagnostic relationships that reflect real-world coding challenges. Additionally, exploring the performance of LLMs on multi-label classification tasks, where a single discharge summary may correspond to multiple ICD-10 codes, could enhance the generalizability of the models in handling overlapping clinical patterns, serving as a continuation of previous research in clinical document classification \cite{922}. Further improvements could include fine tuning the models with domain specific data to enhance their understanding of clinical language and improve performance on nuanced cases. Moreover, conducting experiments with larger datasets and incorporating real-world clinical notes from multiple healthcare institutions could improve model robustness and generalizability. Comparing the performance of ensemble models that combine outputs from both reasoning and non-reasoning models may also provide a more balanced approach to complex classification tasks, leveraging the strengths of both model types to enhance coding accuracy and consistency.

\section{Conclusion}\label{Sec_Conclusion}
The comprehensive evaluation of reasoning and non-reasoning LLMs in classifying ICD-10 coded discharge summaries provides key insights into model performance and consistency. The results reveal that reasoning models outperformed non-reasoning models in terms of overall accuracy and F1 scores, with reasoning models achieving an average accuracy of 71\% and a macro F1 score of 67\%, compared to 68\% and 60\%, respectively, for non-reasoning models. However, non-reasoning models demonstrated higher consistency, averaging 91\% compared to 84\% for reasoning models, suggesting greater reliability across repeated trials. These results indicate that while reasoning models exhibit strength in handling complex clinical cases and achieving higher classification accuracy, non-reasoning models provide more stable and consistent performance in structured ICD-10 classification tasks.

\clearpage

\bibliographystyle{elsarticle-num} 
\bibliography{ref.bib}% common bib file

\begin{thebibliography}{10}
\expandafter\ifx\csname url\endcsname\relax
  \def\url#1{\texttt{#1}}\fi
\expandafter\ifx\csname urlprefix\endcsname\relax\def\urlprefix{URL }\fi
\expandafter\ifx\csname href\endcsname\relax
  \def\href#1#2{#2} \def\path#1{#1}\fi

\bibitem{963}
R.~Nester, Medical coding market size \& share, growth trends 2037,
  \url{https://www.researchnester.com/reports/medical-coding-market/5899},
  accessed: 2025-03-16 (2024).

\bibitem{390}
G.~K. Savova, J.~J. Masanz, P.~V. Ogren, J.~Zheng, S.~Sohn, K.~C.
  Kipper-Schuler, C.~G. Chute, Mayo clinical text analysis and knowledge
  extraction system (ctakes): architecture, component evaluation and
  applications, Journal of the American Medical Informatics Association 17~(5)
  (2010) 507--513.

\bibitem{965}
C.~Gao, M.~Goswami, J.~Chen, A.~Dubrawski, Classifying unstructured clinical
  notes via automatic weak supervision, in: Machine Learning for Healthcare
  Conference, PMLR, 2022, pp. 673--690.

\bibitem{977}
H.~Meyer, Coding complexity: Us health care gets ready for the coming of
  icd-10, Health Affairs 30~(5) (2011) 968--974.

\bibitem{978}
A.~Atutxa, A.~D. de~Ilarraza, K.~Gojenola, M.~Oronoz, O.~Perez-de Vi{\~n}aspre,
  Interpretable deep learning to map diagnostic texts to icd-10 codes,
  International journal of medical informatics 129 (2019) 49--59.

\bibitem{964}
A.~Sammani, A.~Bagheri, P.~G. van~der Heijden, A.~S. Te~Riele, A.~F. Baas,
  C.~Oosters, D.~Oberski, F.~W. Asselbergs, Automatic multilabel detection of
  icd10 codes in dutch cardiology discharge letters using neural networks, NPJ
  digital medicine 4~(1) (2021) 37.

\bibitem{979}
A.~Mclaughlin, J.~Hardt, J.~Canavan, M.~Donnelly, Diagnosis-related group-based
  reimbursement is unrealistic for icus, Critical Care 13~(Suppl 1) (2009)
  P485.

\bibitem{980}
S.~V. Kusnoor, M.~N. Blasingame, A.~M. Williams, S.~J. DesAutels, J.~Su, N.~B.
  Giuse, A narrative review of the impact of the transition to icd-10 and
  icd-10-cm/pcs, JAMIA open 3~(1) (2020) 126--131.

\bibitem{981}
A.~Mustafa, U.~Naseem, M.~R. Azghadi, Large language models vs human for
  classifying clinical documents, International Journal of Medical Informatics
  (2025) 105800.

\bibitem{966}
Z.~Nazi, W.~Peng, Large language models in healthcare and medical domain: A
  review. informatics, 11, 57 (2024).

\bibitem{967}
B.~Mesk{\'o}, The impact of multimodal large language models on health care’s
  future, Journal of medical Internet research 25 (2023) e52865.

\bibitem{968}
A.~Far, A.~Bastani, A.~Lee, O.~Gologorskaya, C.-Y. Huang, M.~J. Pletcher, J.~C.
  Lai, J.~Ge, Evaluating the positive predictive value of code-based
  identification of cirrhosis and its complications utilizing gpt-4, Hepatology
   10--1097.

\bibitem{969}
T.~Kwon, K.~T.-i. Ong, D.~Kang, S.~Moon, J.~R. Lee, D.~Hwang, B.~Sohn, Y.~Sim,
  D.~Lee, J.~Yeo, Large language models are clinical reasoners: Reasoning-aware
  diagnosis framework with prompt-generated rationales, in: Proceedings of the
  AAAI conference on artificial intelligence, Vol.~38, 2024, pp. 18417--18425.

\bibitem{970}
J.~Miao, C.~Thongprayoon, S.~Suppadungsuk, P.~Krisanapan, Y.~Radhakrishnan,
  W.~Cheungpasitporn, Chain of thought utilization in large language models and
  application in nephrology, Medicina 60~(1) (2024) 148.

\bibitem{971}
S.~Bhatia, Next-generation healthcare information systems: Integrating
  chain-of-thought reasoning and adaptive retrieval in large-scale document
  analysis, Available at SSRN 5029973 (2024).

\bibitem{377}
H.~Shi, P.~Xie, Z.~Hu, M.~Zhang, E.~P. Xing, Towards automated icd coding using
  deep learning, arXiv preprint arXiv:1711.04075 (2017).

\bibitem{433}
S.~Gehrmann, F.~Dernoncourt, Y.~Li, E.~T. Carlson, J.~T. Wu, J.~Welt,
  J.~Foote~Jr, E.~Moseley, D.~W. Grant, P.~D. Tyler, et~al., A comparison of
  rule-based and deep learning models for patient phenotyping, preprint, arxiv.
  org (2017).

\bibitem{426}
A.~Karmakar, Classifying medical notes into standard disease codes using
  machine learning, arXiv preprint arXiv:1802.00382 (2018).

\bibitem{958}
R.~Li, X.~Wang, H.~Yu, Exploring llm multi-agents for icd coding, arXiv
  preprint arXiv:2406.15363 (2024).

\bibitem{922}
A.~Mustafa, M.~Rahimi~Azghadi, Clustered automated machine learning (caml)
  model for clinical coding multi-label classification, International Journal
  of Machine Learning and Cybernetics (2024) 1--23.

\end{thebibliography}

\clearpage

\begin{appendices}
\section{Appendix}
\subsection{Tables}

%% Reasoning Models
\begin{table}[h]
\caption{A41 Accuracy, F1 score \& Consistency by LLM}
\label{table_A41}
\begin{tabular}{@{}L{140pt}C{80pt}C{80pt}C{80pt} @{}}
\toprule

\textbf{\makecell{Model}}
&\textbf{\makecell{Accuracy}}
&\textbf{\makecell{F1 Score}}
&\textbf{\makecell{Consistency}}   \\

\midrule
		
Deepseek Reasoner		&\textbf{87.00\%}	&86.22\%	&\textbf{96.67\%} \\
Gemini 2 Flash Thinking	&86.67\%	&\textbf{87.34\%}	&85.67\% \\
GPT 3o Mini				&86.00\%	&85.11\%	&96.33\% \\
Qwen QWQ				&76.00\%	&71.43\%	&78.00\% \\
\midrule
Deepseek Chat			&86.67\%	&86.39\%	&93.67\% \\
Gemini 2 Flash			&76.33\%	&74.18\%	&94.00\% \\
GPT 4o Mini				&71.67\%	&65.86\%	&78.00\% \\
Llama 3.3				&71.00\%	&62.66\%	&85.67\% \\

\bottomrule
\end{tabular}

\end{table}

%% Reasoning Models
\begin{table}[h]
\caption{E78 Accuracy, F1 score \& Consistency by LLM}
\label{table_E78}
\begin{tabular}{@{}L{140pt}C{80pt}C{80pt}C{80pt} @{}}
\toprule

\textbf{\makecell{Model}}
&\textbf{\makecell{Accuracy}}
&\textbf{\makecell{F1 Score}}
&\textbf{\makecell{Consistency}}   \\

\midrule
		
Deepseek Reasoner		&\textbf{80.67\%}	&\textbf{78.52\%}	&98.33\% \\
Gemini 2 Flash Thinking	&80.00\%	&78.26\%	&97.00\% \\
GPT 3o Mini				&80.33\%	&78.23\%	&\textbf{98.67\%} \\
Qwen QWQ				&69.33\%	&60.00\%	&79.33\% \\
\midrule
Deepseek Chat			&64.33\%	&47.80\%	&92.67\% \\
Gemini 2 Flash			&74.00\%	&73.83\%	&91.33\% \\
GPT 4o Mini				&56.33\%	&23.39\%	&91.00\% \\
Llama 3.3				&70.67\%	&72.33\%	&74.33\% \\

\bottomrule
\end{tabular}
\end{table}

%% Reasoning Models
\begin{table}[h]
\caption{I10 Accuracy, F1 score \& Consistency by LLM}
\label{table_I10}
\begin{tabular}{@{}L{140pt}C{80pt}C{80pt}C{80pt} @{}}
\toprule

\textbf{\makecell{Model}}
&\textbf{\makecell{Accuracy}}
&\textbf{\makecell{F1 Score}}
&\textbf{\makecell{Consistency}}   \\

\midrule

Deepseek Reasoner		&82.67\%	&84.88\%	&99.00\%  \\
Gemini 2 Flash Thinking	&\textbf{83.00\%}	&\textbf{85.22\%}	&99.33\%  \\
GPT 3o Mini				&\textbf{83.00\%}	&\textbf{85.22\%}	&\textbf{100.00\%} \\
Qwen QWQ				&75.67\%	&75.75\%	&67.00\%  \\
\midrule
Deepseek Chat			&77.33\%	&76.55\%	&93.67\%  \\
Gemini 2 Flash			&81.67\%	&83.58\%	&96.67\%  \\
GPT 4o Mini				&69.67\%	&65.66\%	&82.33\%  \\
Llama 3.3				&77.00\%	&77.08\%	&87.67\%  \\

\bottomrule
\end{tabular}
\end{table}

%% Reasoning Models
\begin{table}[h]
\caption{I13 Accuracy, F1 score \& Consistency by LLM}
\label{table_I13}
\begin{tabular}{@{}L{140pt}C{80pt}C{80pt}C{80pt} @{}}
\toprule

\textbf{\makecell{Model}}
&\textbf{\makecell{Accuracy}}
&\textbf{\makecell{F1 Score}}
&\textbf{\makecell{Consistency}}   \\

\midrule

Deepseek Reasoner		&70.33\%	&58.60\%	&75.00\% \\
Gemini 2 Flash Thinking	&\textbf{84.33\%}	&\textbf{86.30\%}	&87.33\% \\
GPT 3o Mini				&82.33\%	&80.59\%	&63.67\% \\
Qwen QWQ				&62.33\%	&40.21\%	&67.67\% \\
\midrule
Deepseek Chat			&83.33\%	&80.31\%	&89.67\% \\
Gemini 2 Flash			&\textbf{84.33\%}	&85.89\%	&\textbf{91.67\%} \\
GPT 4o Mini				&58.33\%	&30.17\%	&80.00\% \\
Llama 3.3				&80.67\%	&79.14\%	&77.33\% \\

\bottomrule
\end{tabular}
\end{table}

%% Reasoning Models
\begin{table}[h]
\caption{I21 Accuracy, F1 score \& Consistency by LLM}
\label{table_I21}
\begin{tabular}{@{}L{140pt}C{80pt}C{80pt}C{80pt} @{}}
\toprule

\textbf{\makecell{Model}}
&\textbf{\makecell{Accuracy}}
&\textbf{\makecell{F1 Score}}
&\textbf{\makecell{Consistency}}   \\

\midrule

Deepseek Reasoner		&86.67\%	&84.73\%	&95.00\% \\
Gemini 2 Flash Thinking	&86.67\%	&85.07\%	&97.33\% \\
GPT 3o Mini				&86.67\%	&84.85\%	&\textbf{98.33\%} \\
Qwen QWQ				&86.00\%	&83.97\%	&89.33\% \\
\midrule
Deepseek Chat			&88.33\%	&86.99\%	&93.00\% \\
Gemini 2 Flash			&85.00\%	&82.76\%	&\textbf{98.33\%} \\
GPT 4o Mini				&80.33\%	&76.31\%	&85.67\% \\
Llama 3.3				&\textbf{95.00\%}	&\textbf{94.98\%}	&94.00\% \\

\bottomrule
\end{tabular}
\end{table}

%% Reasoning Models
\begin{table}[h]
\caption{I25 Accuracy, F1 score \& Consistency by LLM}
\label{table_I25}
\begin{tabular}{@{}L{140pt}C{80pt}C{80pt}C{80pt} @{}}
\toprule

\textbf{\makecell{Model}}
&\textbf{\makecell{Accuracy}}
&\textbf{\makecell{F1 Score}}
&\textbf{\makecell{Consistency}}   \\

\midrule

Deepseek Reasoner		&93.67\%	&94.04\%	&\textbf{99.67\%} \\
Gemini 2 Flash Thinking	&92.00\%	&92.59\%	&96.33\% \\
GPT 3o Mini				&94.00\%	&94.34\%	&\textbf{99.67\%} \\
Qwen QWQ				&94.00\%	&94.34\%	&95.67\% \\
\midrule
Deepseek Chat			&95.33\%	&95.54\%	&99.00\% \\
Gemini 2 Flash			&90.67\%	&91.46\%	&97.00\% \\
GPT 4o Mini				&\textbf{96.00\%}	&\textbf{96.08\%}	&92.67\% \\
Llama 3.3				&87.33\%	&88.76\%	&91.00\% \\

\bottomrule
\end{tabular}
\end{table}

%% Reasoning Models
\begin{table}[h]
\caption{Y92 Accuracy, F1 score \& Consistency by LLM}
\label{table_Y92}
\begin{tabular}{@{}L{140pt}C{80pt}C{80pt}C{80pt} @{}}
\toprule

\textbf{\makecell{Model}}
&\textbf{\makecell{Accuracy}}
&\textbf{\makecell{F1 Score}}
&\textbf{\makecell{Consistency}}   \\

\midrule

Deepseek Reasoner		&50.00\%	&2.60\%		&94.33\%  \\
Gemini 2 Flash Thinking	&\textbf{51.33\%}	&7.59\%		&86.67\%  \\
GPT 3o Mini				&50.00\%	&0.00\%		&\textbf{100.00\%} \\
Qwen QWQ				&50.00\%	&0.00\%		&98.67\%  \\
\midrule
Deepseek Chat			&50.00\%	&0.00\%		&\textbf{100.00\%} \\
Gemini 2 Flash			&51.00\%	&\textbf{19.67\%}	&91.67\%  \\
GPT 4o Mini				&50.00\%	&0.00\%		&\textbf{100.00\%} \\
Llama 3.3				&50.00\%	&0.00\%		&99.67\%  \\

\bottomrule
\end{tabular}
\end{table}

%% Reasoning Models
\begin{table}[h]
\caption{Z51 Accuracy, F1 score \& Consistency by LLM}
\label{table_Z51}
\begin{tabular}{@{}L{140pt}C{80pt}C{80pt}C{80pt} @{}}
\toprule

\textbf{\makecell{Model}}
&\textbf{\makecell{Accuracy}}
&\textbf{\makecell{F1 Score}}
&\textbf{\makecell{Consistency}}   \\

\midrule

Deepseek Reasoner		&85.00\%	&83.27\%	&72.00\% \\
Gemini 2 Flash Thinking	&\textbf{85.33\%}	&\textbf{86.83\%}	&65.67\% \\
GPT 3o Mini				&51.33\%	&6.41\%		&98.00\% \\
Qwen QWQ				&46.33\%	&3.59\%		&80.33\% \\
\midrule
Deepseek Chat			&51.00\%	&3.92\%		&\textbf{98.67\%} \\
Gemini 2 Flash			&70.67\%	&63.93\%	&83.33\% \\
GPT 4o Mini				&55.33\%	&19.28\%	&83.33\% \\
Llama 3.3				&52.33\%	&8.92\%		&90.00\% \\

\bottomrule
\end{tabular}
\end{table}

%% Reasoning Models
\begin{table}[h]
\caption{Z79 Accuracy, F1 score \& Consistency by LLM}
\label{table_Z79}
\begin{tabular}{@{}L{140pt}C{80pt}C{80pt}C{80pt} @{}}
\toprule

\textbf{\makecell{Model}}
&\textbf{\makecell{Accuracy}}
&\textbf{\makecell{F1 Score}}
&\textbf{\makecell{Consistency}}   \\

\midrule

Deepseek Reasoner		&\textbf{55.00\%}	&51.61\%	&39.67\%  \\
Gemini 2 Flash Thinking	&54.00\%	&\textbf{57.14\%}	&32.67\%  \\
GPT 3o Mini				&50.00\%	&0.00\%		&\textbf{100.00\%} \\
Qwen QWQ				&49.33\%	&3.80\%		&83.67\%  \\
\midrule
Deepseek Chat			&51.67\%	&7.64\%		&94.00\%  \\
Gemini 2 Flash			&52.33\%	&36.44\%	&85.67\%  \\
GPT 4o Mini				&50.00\%	&2.60\%		&96.00\%  \\
Llama 3.3				&50.33\%	&2.61\%		&93.00\%  \\

\bottomrule
\end{tabular}
\end{table}

%% Reasoning Models
\begin{table}[h]
\caption{Z87 Accuracy, F1 score \& Consistency by LLM}
\label{table_Z87}
\begin{tabular}{@{}L{140pt}C{80pt}C{80pt}C{80pt} @{}}
\toprule

\textbf{\makecell{Model}}
&\textbf{\makecell{Accuracy}}
&\textbf{\makecell{F1 Score}}
&\textbf{\makecell{Consistency}}   \\

\midrule

Deepseek Reasoner		&47.33\%	&26.17\%	&58.67\%  \\
Gemini 2 Flash Thinking	&49.67\%	&55.46\%	&36.33\%  \\
GPT 3o Mini				&50.00\%	&0.00\%		&\textbf{100.00\%} \\
Qwen QWQ				&43.33\%	&39.72\%	&41.00\%  \\
\midrule
Deepseek Chat			&50.00\%	&0.00\%		&99.67\%  \\
Gemini 2 Flash			&\textbf{53.33\%}	&\textbf{62.37\%}	&77.00\%  \\
GPT 4o Mini				&50.00\%	&0.00\%		&98.33\%  \\
Llama 3.3				&49.00\%	&2.55\%		&90.33\%  \\

\bottomrule
\end{tabular}
\end{table}

\clearpage

\subsection{Charts}

\begin{figure}[h]
\centering
\includegraphics[width=1.0\textwidth]{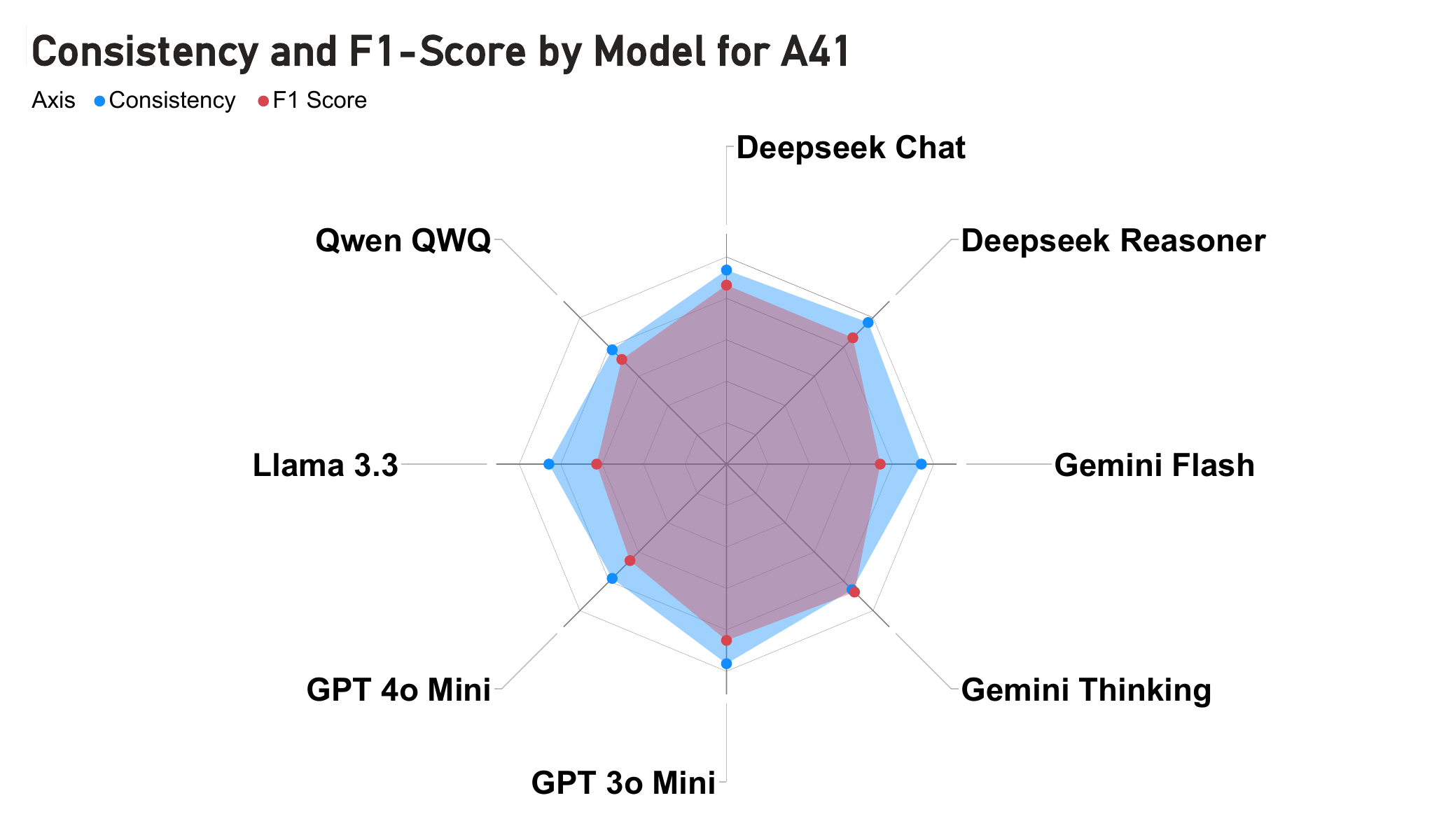}
\noindent
\caption{Consistency and F1 Score by Model for A41.}
\label{g_Rad_A41}
\end{figure}

\begin{figure}[h]
\centering
\includegraphics[width=1.0\textwidth]{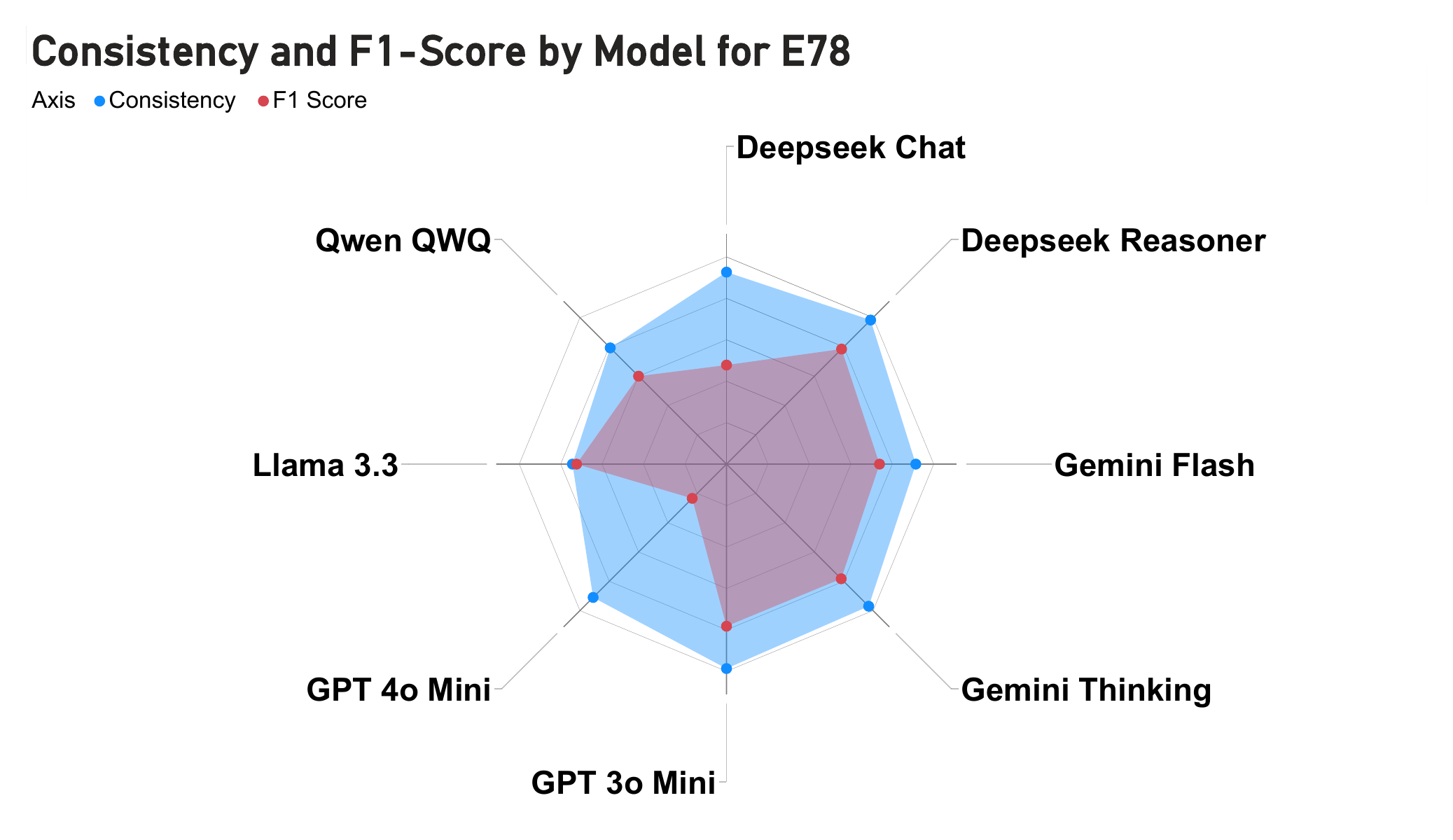}
\noindent
\caption{Consistency and F1 Score by Model for E78.}
\label{g_Rad_E78}
\end{figure}

\begin{figure}[h]
\centering
\includegraphics[width=1.0\textwidth]{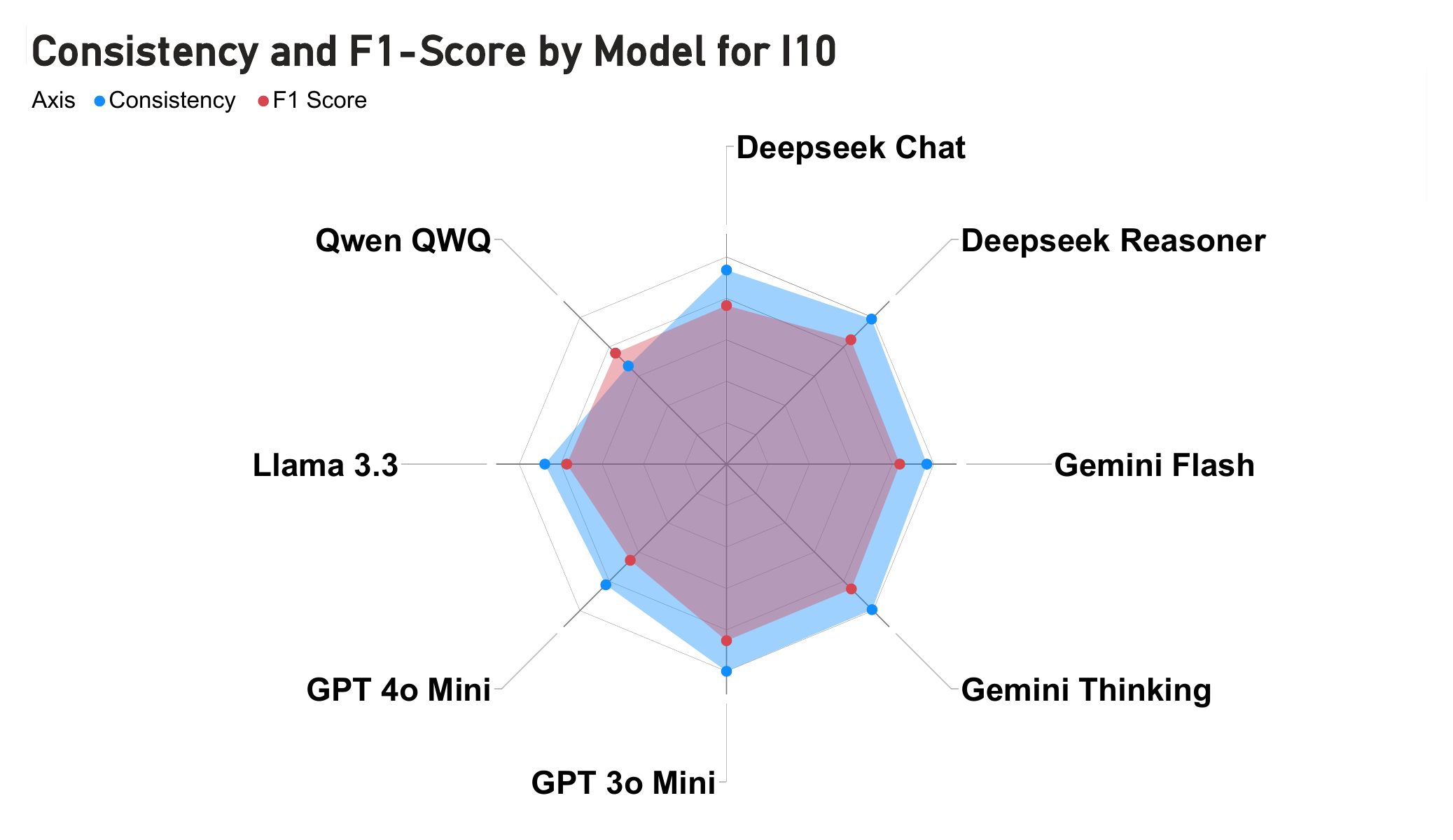}
\noindent
\caption{Consistency and F1 Score by Model for I10.}
\label{g_Rad_i10}
\end{figure}

\begin{figure}[h]
\centering
\includegraphics[width=1.0\textwidth]{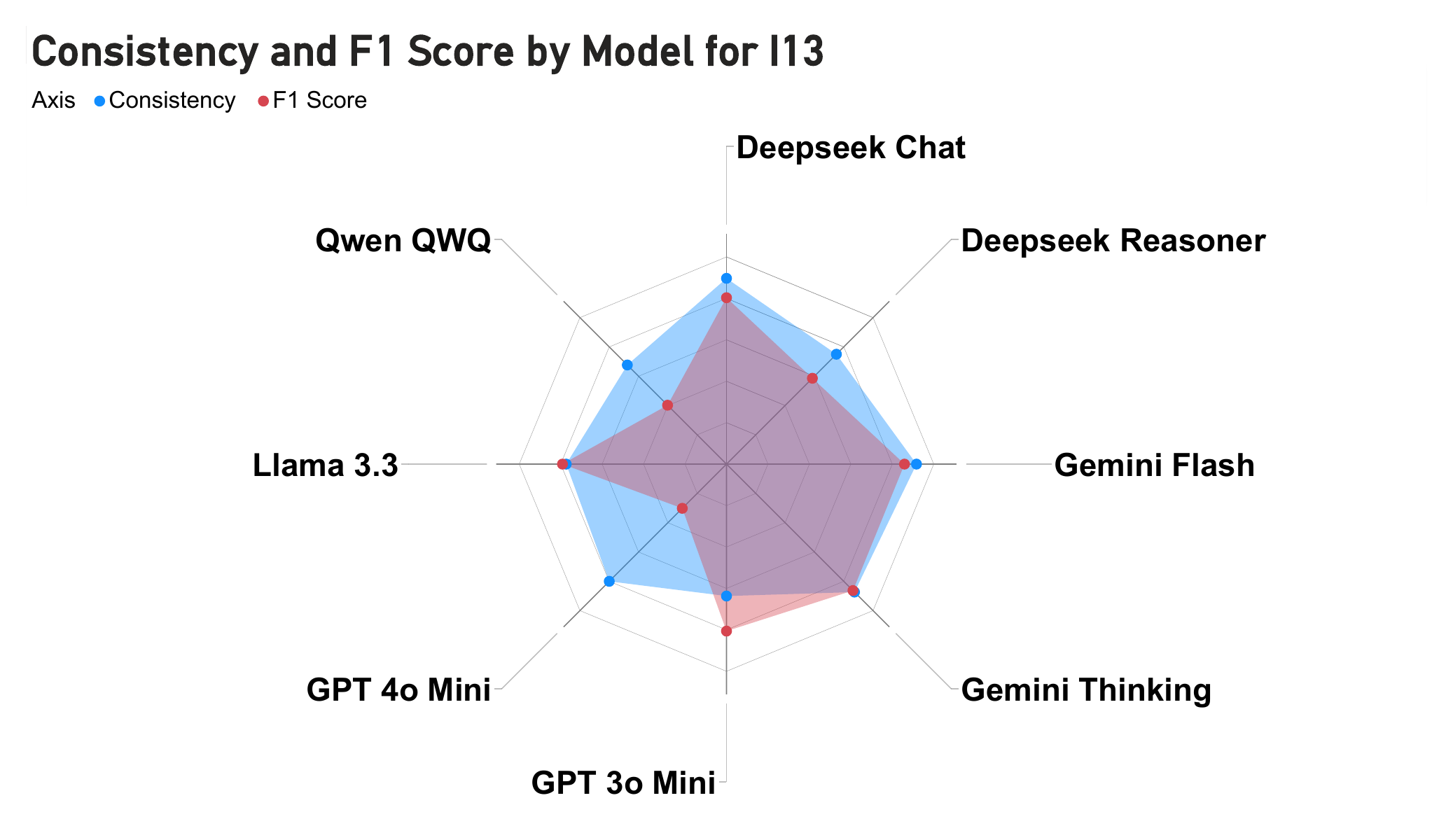}
\noindent
\caption{Consistency and F1 Score by Model for I13.}
\label{g_Rad_i13}
\end{figure}

\begin{figure}[h]
\centering
\includegraphics[width=1.0\textwidth]{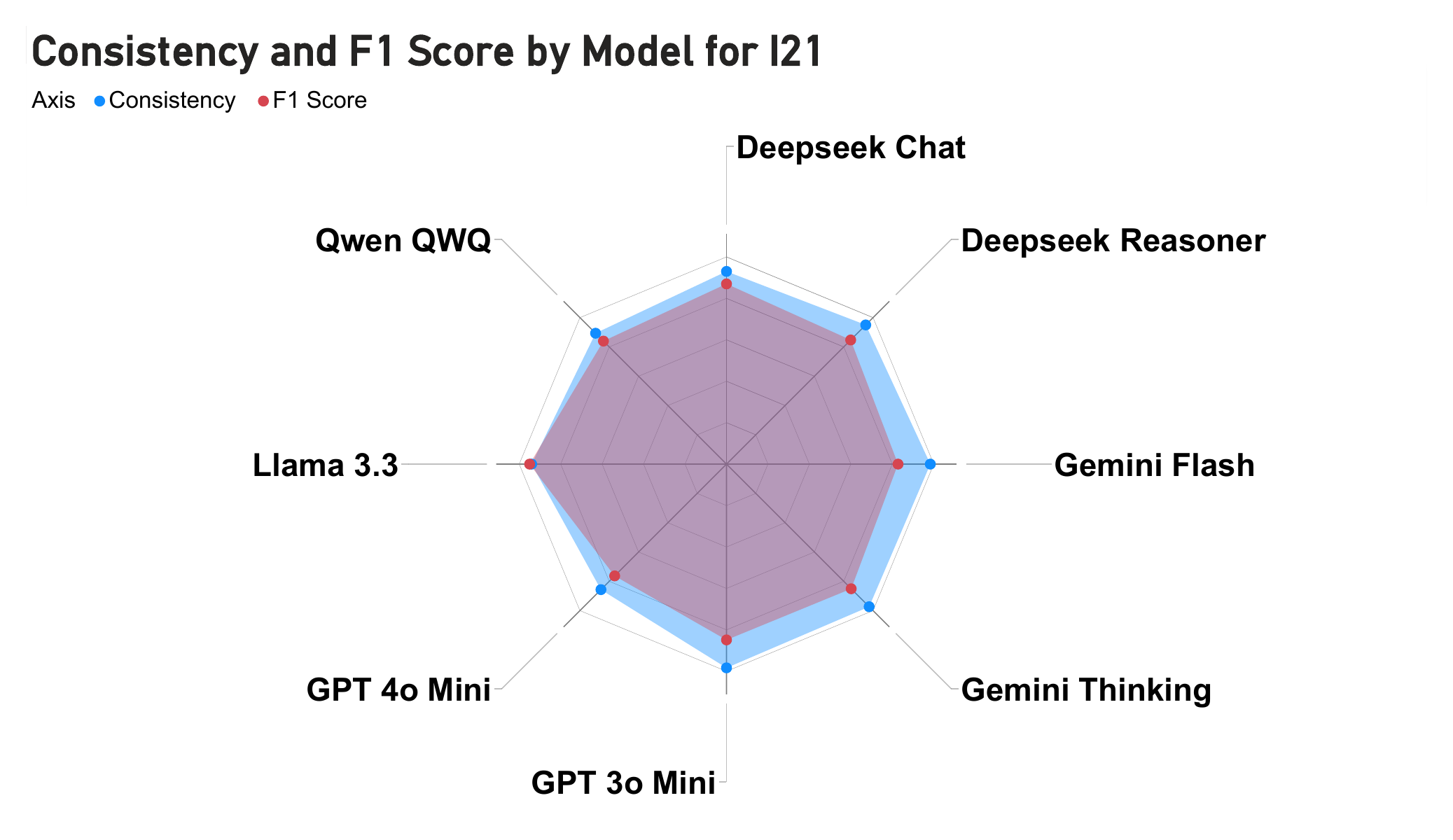}
\noindent
\caption{Consistency and F1 Score by Model for I21.}
\label{g_Rad_i21}
\end{figure}

\begin{figure}[h]
\centering
\includegraphics[width=1.0\textwidth]{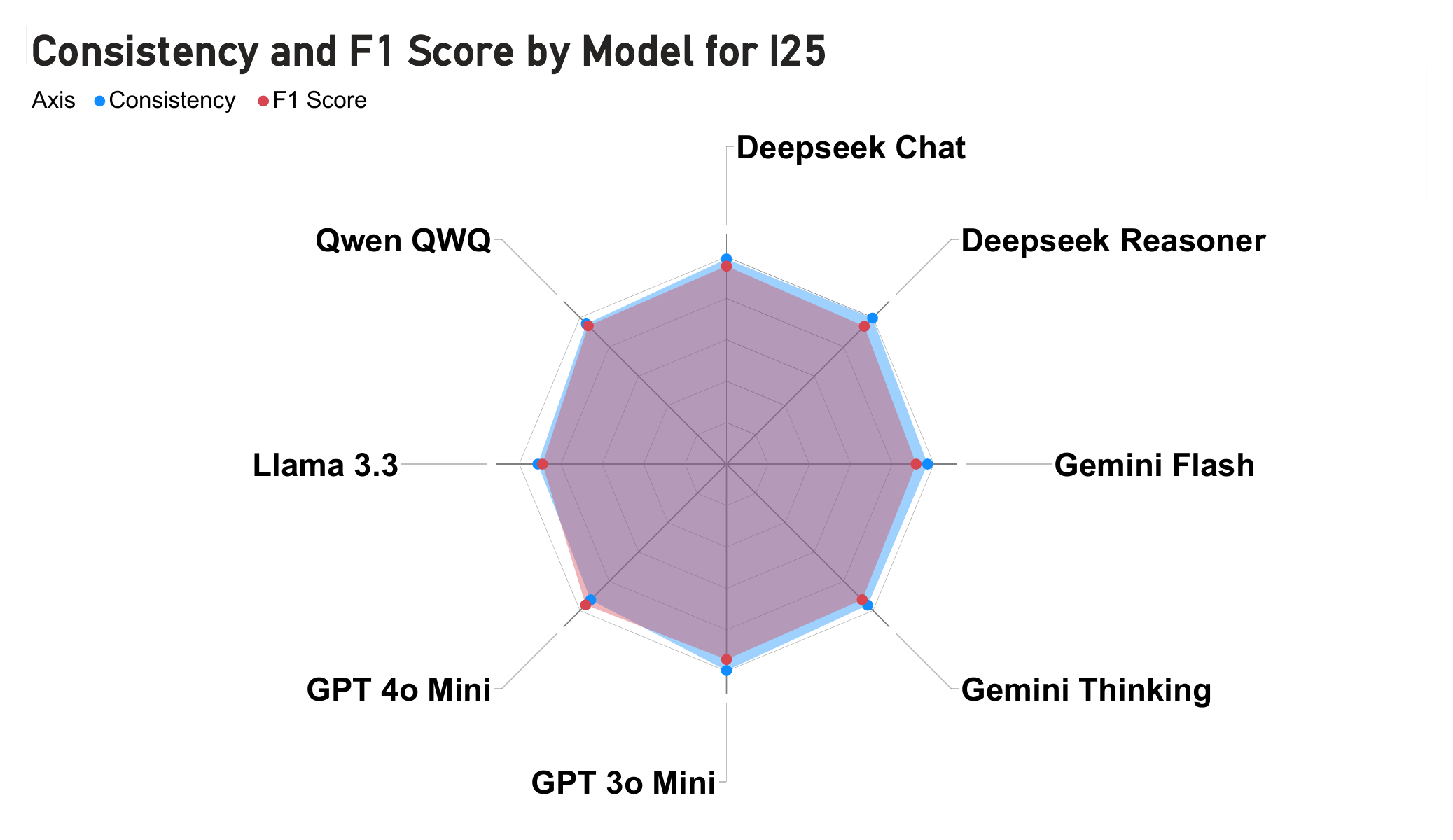}
\noindent
\caption{Consistency and F1 Score by Model for I25.}
\label{g_Rad_i25}
\end{figure}

\begin{figure}[h]
\centering
\includegraphics[width=1.0\textwidth]{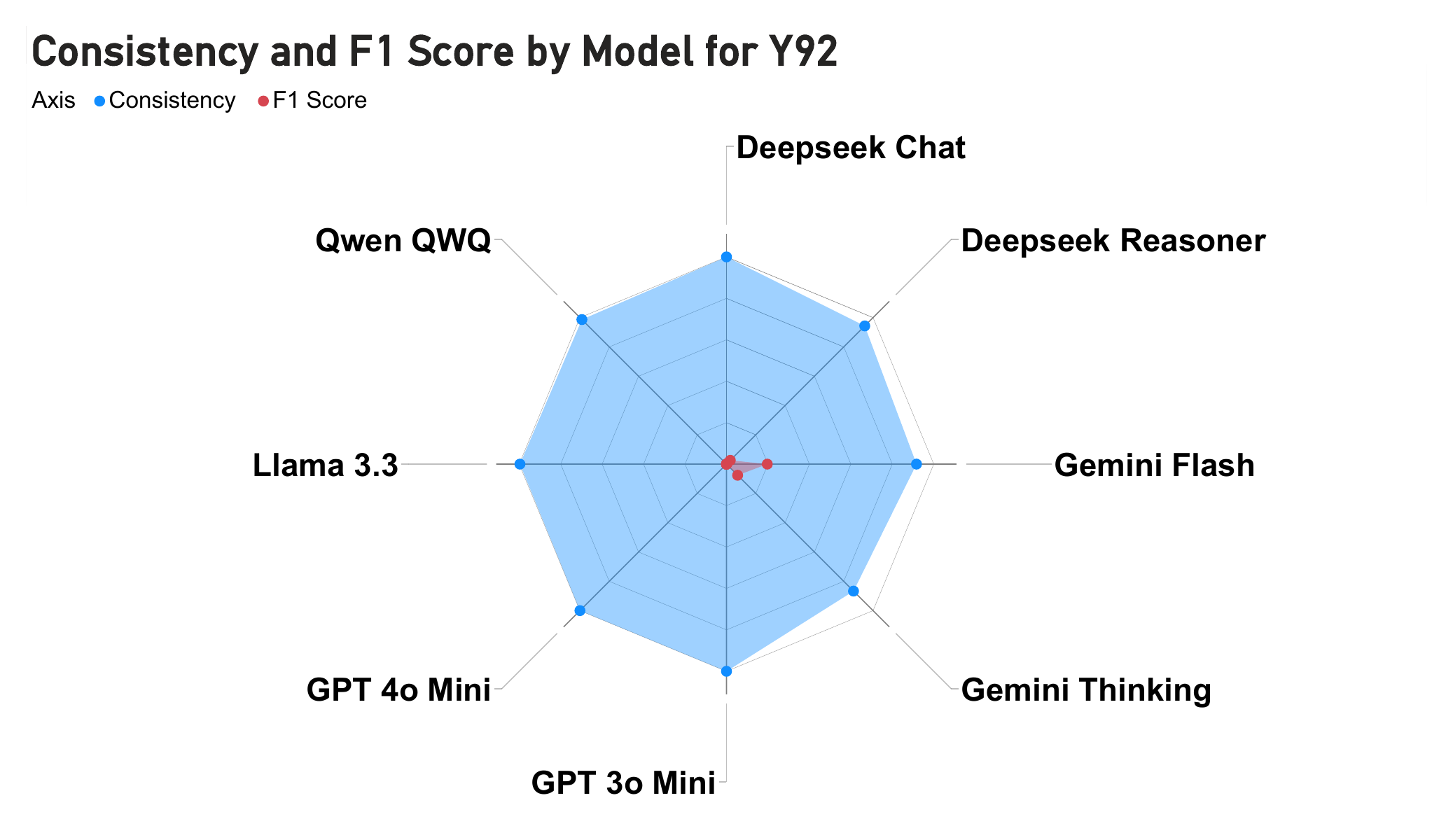}
\noindent
\caption{Consistency and F1 Score by Model for Y92.}
\label{g_Rad_Y92}
\end{figure}

\begin{figure}[h]
\centering
\includegraphics[width=1.0\textwidth]{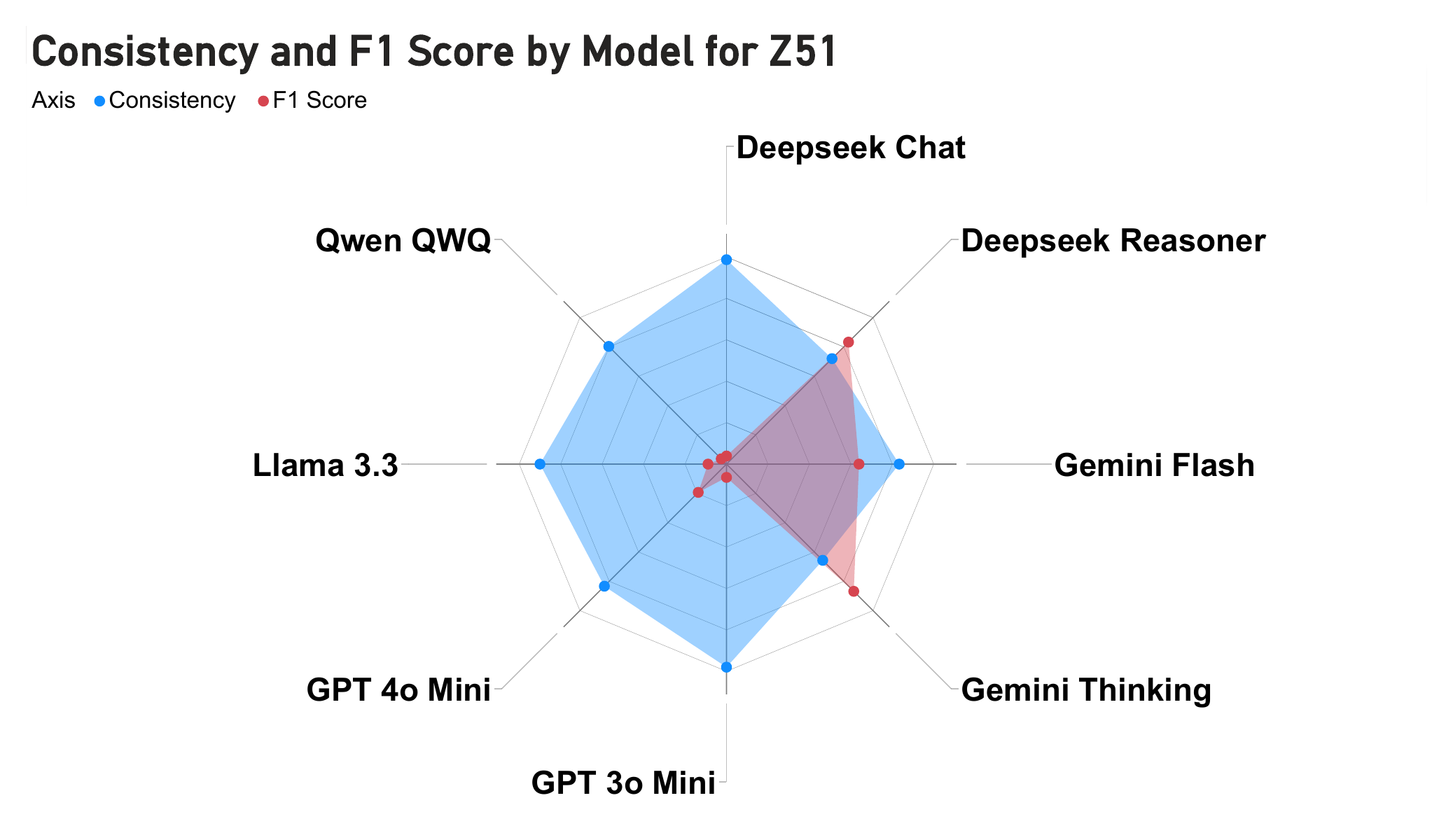}
\noindent
\caption{Consistency and F1 Score by Model for Z51.}
\label{g_Rad_Z51}
\end{figure}

\begin{figure}[h]
\centering
\includegraphics[width=1.0\textwidth]{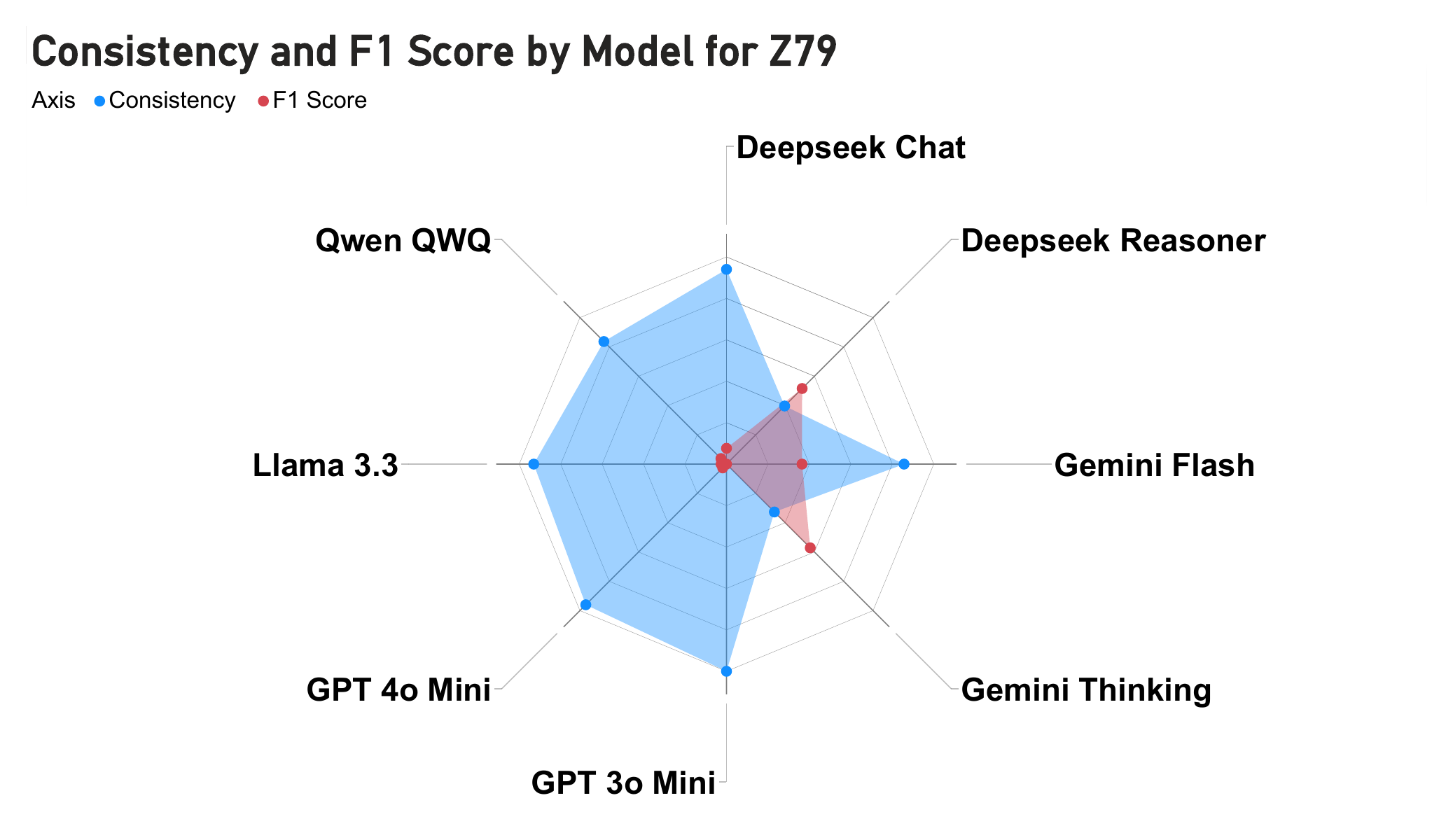}
\noindent
\caption{Consistency and F1 Score by Model for Z79.}
\label{g_Rad_Z79}
\end{figure}

\begin{figure}[h]
\centering
\includegraphics[width=1.0\textwidth]{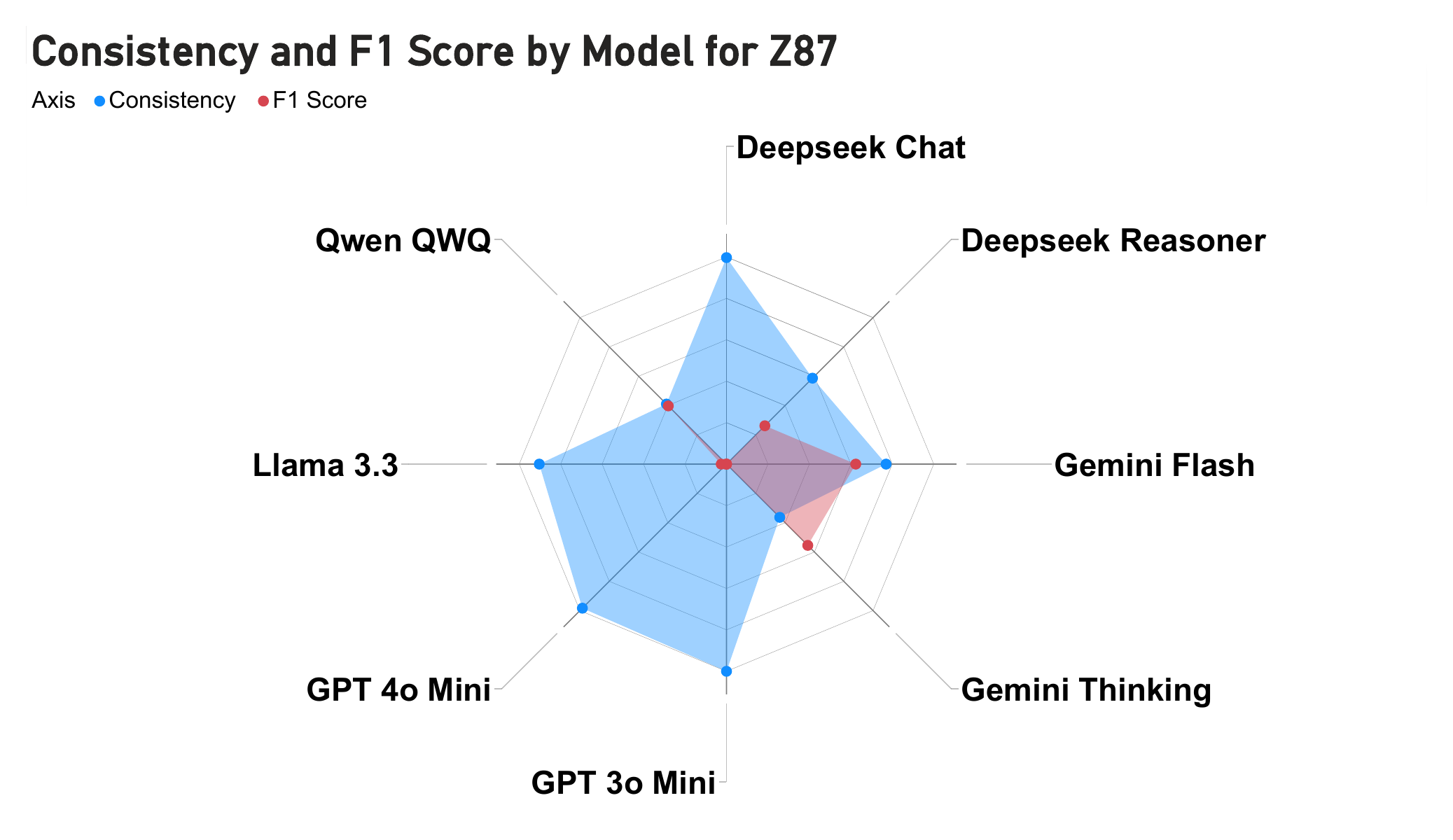}
\noindent
\caption{Consistency and F1 Score by Model for Z87.}
\label{g_Rad_Z87}
\end{figure}

\end{appendices}

\end{document}